\documentclass{article}

\usepackage{arxiv}

\usepackage[utf8]{inputenc} 
\usepackage[T1]{fontenc}    
\usepackage{hyperref}       
\usepackage{url}            
\usepackage{booktabs}       
\usepackage{amsfonts}       
\usepackage{nicefrac}       
\usepackage{microtype}      
\usepackage{lipsum}
\usepackage{cite}
\usepackage{graphicx}
\usepackage{bm}
\usepackage[ruled,linesnumbered]{algorithm2e}
\usepackage{enumitem}

\SetKwInput{KwInput}{Input}     
\SetKwInput{KwOutput}{Output}   

\title{Adaptive Histogram-Based Gradient Boosted Trees for Federated Learning}
\author{
    Yuya Jeremy Ong \\
    IBM Research - Almaden \\
    San Jose, CA 95120 \\
    \texttt{yuyajong@ibm.com} \\
    \And
    Yi Zhou \\
    IBM Research - Almaden \\
    San Jose, CA 95120 \\
    \texttt{yi.zhou@ibm.com} \\
    \And
    Nathalie Baracaldo \\
    IBM Research - Almaden \\
    San Jose, CA 95120 \\
    \texttt{baracald@us.ibm.com} \\
    \And
    Heiko Ludwig \\
    IBM Research - Almaden \\
    San Jose, CA 95120 \\
    \texttt{hludwig@us.ibm.com} \\
}

\begin{document}
\maketitle

\begin{abstract}
Federated Learning (FL) is an approach to collaboratively train a model across multiple parties without sharing data between parties or an aggregator. It is used both in the consumer domain to protect personal data as well as in enterprise settings, where dealing with data domicile regulation and the pragmatics of data silos are the main drivers. While gradient boosted tree implementations such as XGBoost have been very successful for many use cases, its federated learning adaptations tend to be very slow due to using cryptographic and privacy methods and have not experienced widespread use. We propose the Party-Adaptive XGBoost (PAX) for federated learning, a novel implementation of gradient boosting which utilizes a party adaptive histogram aggregation method, without the need for data encryption. It constructs a surrogate representation of the data distribution for finding splits of the decision tree. Our experimental results demonstrate strong model performance, especially on non-IID distributions, and significantly faster training run-time across different data sets than existing federated implementations. This approach makes the use of gradient boosted trees practical in enterprise federated learning.
\end{abstract}

\section{Introduction}
Gradient boosted decision trees, e.g., XGBoost \cite{chen2016xgboost}, have recently been proposed as a new ensemble tree-based model to improve the performance of CART decision trees.
It utilizes a gradient-boosting based approach to optimize the tree split against a predefined loss function, e.g., mean-squared loss for regression problems and cross-entropy loss for classification problems, etc. 
As it exhibits good performances for different types of machine learning (ML) problems, it has gained a lot of attention in both the Kaggle community and real-world applications.
Moreover, since it is a tree-based model, it also pertains good explainability of tree models which deep learning models struggle with.
	
Federated learning (FL) has been recently proposed \cite{mcmahan2016communication} to enable collaborative training of ML models among multiple parties where each party can keep its data to itself and there is no need to bring data to a central location. 
This approach has been beneficial for use cases in which the privacy of consumer data is important, such as data on cell phone apps, as well as scenarios in which regulation limits the movement of data. For example, many countries have adopted legislation that person identifying data cannot be moved outside of their jurisdiction or a jurisdiction with lesser privacy protection. 
Examples of such regulations include the European General Data Protection Regulation (GDPR), the California Consumer Privacy Act (CCPA), and the Health Insurance Portability and Accountability Act (HIPAA). 
Federated learning also found good use in scenarios where large amounts of data are produced in different locations such as different data centers, application silos, or Cloud providers, where bringing all data together would be impractical. Applications of federated learning on consumer devices, such as mobile devices or vehicles, typically require advanced privacy guarantees such as those provided by differential privacy and the addition of differentially private noise as part of the learning process. 
Different frameworks for FL have been proposed, including FATE \cite{fate:2020}, IBM FL \cite{ludwig2020ibm} and Tensorflow Federated \cite{tff:2020}, both for production and experimental use. 
	
The most common topology for federated learning uses a single aggregator (or coordinator), which coordinates the federated learning process. The learning process is divided into one part that takes place at each party to the federated learning job and another that takes place at the aggregator. When designing a federated learning algorithm we have to choose what goes where. There has been significant work on training neural networks, and other models based on a gradient descent training process, in a federated way. In this case, the local parties train neural networks and send updates on their locally trained models - weights or gradients - to the aggregator, which in turn merges the local updates and creates a new shared model that is often sent back to the parties for another round of training off this updated model. 
	
In particular, federated learning poses some specific issues typically not present in centralized machine learning: there is no overall perspective of the training data and its properties since it is not shared. This limits the assumptions we can make about the training dataset. The data also might be unbalanced between parties. One party might have many more samples than others and the distributions might be quite different. This is more a feature than a bug as the point of the federation is to bring data together that might reflect local variety and - jointly - represent a more globally representative dataset. However, this may entail distributional imbalances, lack of statistical independence, missing values, and other issues related to the independent and identical distribution (IID) assumption of the dataset. 
	
Comparing to other popular ML models, like, neural networks and logistic models, among others, XGBoost has many unique advantages that increase its suitability for FL systems.
Firstly, XGBoost can deal with missing data out of the box, while other ML models require additional processing on the training data. Popular approaches, such as replacing the missing feature values with the mean value, usually require coordination among parties and therefore result in large communication overhead.
Secondly, XGBoost does not require an extensive set of pre-processing operations, which saves significant communication costs and reduces private data leakage as a result of the additional communication required.
The performance of classic ML models, for instance, linear models, logistic models and SVMs, etc., often relies heavily on good pre-processing strategies. 
And the standard pre-processing methods, like normalization and scaling, again require coordination among parties especially when parties' local datasets are heterogeneous, a well-known characteristic of FL \cite{mcmahan2016communication}. 
	
While some federated approaches to gradient boosting trees have been proposed, they typically focus on consumer FL, where the application of expensive security and privacy methods are necessary. As a result, these approaches often entail poor model performance compared to centralized gradient boosting trees. Adoption has been limited for enterprise settings.
	
In this paper, we focus on enterprise use cases and we propose Party-Adaptive XGBoost (PAX), an algorithm to train XGBoost models in FL settings. We demonstrate that PAX outperforms existing state-of-the-art approaches under varied data distributions that mimic real FL setting. \\
	
\noindent{\bf Our Contributions:}
\begin{itemize}[leftmargin=*]
	\item We design a federated learning approach for XGBoost, PAX, in which parties share model information based on a surrogate histogram representation of their data. The resolution of the histograms are adapted for each party to limit data leakage.
	\item We introduce a novel way to fuse the party-generated histogram based on the downstream learning objective. When dealing with a regression problem, we align the histogram based on the smallest resolution. In contrast, we use a larger resolution for classification problems. 
	\item Our proposed PAX, an algorithm to train XGBoost models in federated learning settings, outperforms existing state-of-the-art approaches even under challenging settings.
	Our experimental results show that the algorithm produces 88\% accuracy even when training participants have data with Non-IID characteristics. We even demonstrate robust results with 91\% accuracy in extremely skewed scenarios, when the number of samples at each participant is significantly different and when each party produces data from different distributions.
\end{itemize}
	
The remainder of this paper is organized as follows. In Section~\ref{sec:related} we briefly discuss the related work. We introduce some preliminary background and notation in Section~\ref{sec:pre}. Our proposed algorithm PAX is presented in Section~\ref{sec:pax}, and we compare its performance with several state-of-the-art approaches and evaluate against different data partition scenarios in Section~\ref{sec:numerical}. We conclude our paper in Section~\ref{sec:conclusion}.

\section{Related Work}\label{sec:related}
Recently, researchers have devoted most of their efforts designing efficient FL algorithms \cite{mcmahan2016communication,konevcny2016federated,li2019convergence} to train neural networks and general linear models. However, we find that relatively small number of works focus on FL algorithms for training gradient boosting models. As of writing, current federated learning frameworks that provide implementations of gradient boosted trees includes FATE \footnote{https://fate.fedai.org/} \cite{fate:2020} and Secure XGBoost from MC2~\footnote{https://github.com/mc2-project/secure-xgboost}.  Largely, the key differences within the existing literature for gradient boosted trees trained via FL are: i) horizontal or vertical FL implementations and ii) use of encryption-based security protocols to protect data.
	
\subsection{Horizontal vs Vertical FL}
Depending on how parties shared the training data samples, FL can be categorized into two different types: horizontal and vertical FL.
Horizontal FL considers the shared set of features, while vertical FL considers the shared set of samples across each party. Depending on the data partition structure, the communication topology of the FL system, as well as the corresponding exchanging information can change largely. 
Many of the existing literature considers a horizontal data partition \cite{liu2019boosting,leung2020towards,yang2019tradeoff,li2019convergence,wang2020cloud,cheng2019secureboost}. Other methods such as SecureBoost \cite{cheng2019secureboost} (also part of FATE\footnote{FATE also provides a horizontal FL variant of GBT}), S-XGB and HESS-XGB \cite{Fang2020AHF}, and SecureGBM \cite{feng2019securegbm} considers the vertical partition of the data. In most of the proposed methods, the information exchanged in these algorithms are based primarily on the gradients, hessians, and tree split decision value thresholds of the loss with respect to each of the party's local data. 
In this work, our proposed algorithm work with horizontal federated learning, which also exchanges gradients, hessians, and also tree split decision value thresholds.
However, our proposed PAX uses a novel party-adaptive histogram-based partition driven approach that we empirically demonstrate can handle Non-IID data better than the baselines.
	
\subsection{Encryption-based protocols}
To address the issue of data privacy and security, various methods have been implemented to prevent an adversary from getting access to the party's local raw data distribution. Many of the existing FL-based gradient boosting methods employ an encryption-based security protocol. For example, SecureBoost \cite{cheng2019secureboost}, FedXGB \cite{liu2019boosting}, and HESS-XGB \cite{Fang2020AHF} employs homomorphic encryption, while methods such as S-XGB \cite{Fang2020AHF} and PrivColl \cite{leung2020towards} use Secret Sharing, and Li et. al \cite{li2020practical} implements a Locality Sensitivity Hashing (LSH) method. 
	
Although majority of the encryption methods provide lossless or near lossless guarantees on privacy protection while maintaining comparable model accuracy than centralized XGBoost, it does so with a significant overhead to both computational and network communication costs. This trade-off, as a result, significantly increases the runtime of the training process. Alternatively, \cite{yang2019tradeoff} proposed a clustering-based k-anonymity scheme, without the use of encryption, for protecting the party's local data. However, a drawback to this approach entails that the size of the virtual cluster needs to be set before hand and does not dynamically adapt to the local party's data distribution, which can be greatly different in a FL system.

In this paper, we focus on enterprise settings where the collaborations do not require secure aggregation schemes. Our design does not utilize cryptosystems, yielding significant training time advantages over some of the prior methods. Our proposed method, instead, utilizes a party-adaptive histogram-based method to approximate the raw distribution of the data through a surrogate representation of the data.
Therefore, parties does not need to reveal their exact local data distributions to others, and they can choose their own granularity to reveal the data information and collaboratively grow the tree.

\section{Preliminaries}\label{sec:pre}
This section introduces some of the preliminaries related to federated learning and centralized gradient boosted trees. We also introduce some of the notation and terminology which we will be referenced throughout the rest of paper.
	
\subsection{Federated Learning}\label{sec:fl}
\begin{figure}[]
	\centering
	\includegraphics[width=0.65\columnwidth]{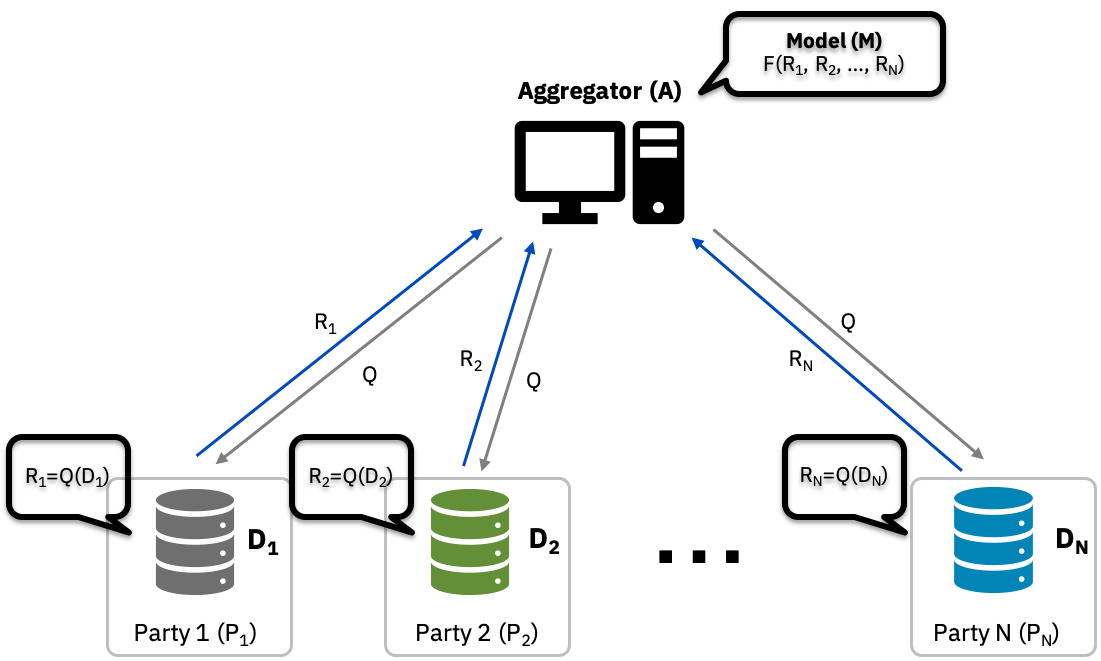}
	\caption{\footnotesize A general architecture of federated learning}
	\label{fig:fl-overview}
\end{figure}

We provide a brief introduction to federated learning and the related terminology in this subsection. Recently, federated learning (FL) has become a popular approach to train machine learning models collaboratively because it does not require transmission or sharing of raw training data to a central place. Throughout the paper, we focus on the type of FL system as shown in Figure~\ref{fig:fl-overview}.
This FL system consists of an aggregator $\mathcal{A}$ and a group of parties $\mathcal{P}_i$, with $i= 1, \dots, n$, which joins FL to collaboratively train a model, usually referred as global model, under the coordination of the aggregator. 
As shown in Figure~\ref{fig:fl-overview}, for each training round, the aggregator issues a query $Q$ to the available parties in the system, a party $\mathcal{P}_i$ upon receiving the query generates a corresponding reply $R_i$ based on its local dataset $\mathcal{D}_i$ and/or its current local ML model and sends the reply back to the aggregator. 
The aggregator then based on the FL algorithm fuses the collected replies and uses the fusion results to update the global ML model.
A FL algorithm may contain several round of training to obtain the final global model.

Most FL algorithms are based on distributed SGD, like FedAvg \cite{mcmahan2016communication,li2019convergence}. They were proposed to train neural networks and can also be applied to general linear models, like logistic models. These FL algorithms assume that a model architecture is provided to both the aggregator and parties before the FL training, and this model architecture, e.g., a neural network structure, won't change during the course of the FL training. Queries in these algorithms usually contains the current global model's parameters and requests the parties' updated model parameters or gradients.
	
However, these existing FL algorithms cannot be applied to train tree based models such as CART decision trees and XGBoost. It is nontrivial to design an algorithm to train a tree based model in a federated learning fashion. Firstly, the model architecture, i.e., the tree structure, will change as the training proceeds. The tree grows as more information is collected. Secondly, it is not straightforward to decide what information will be shared by the parties with the aggregator without revealing the parties' private data. Thirdly, it is not easy to deal with data heterogeneity among parties. Finally, the aggregator should be able to evaluate different termination criteria from a global perspective as trees based models have various termination criteria, e.g., maximum tree depth and minimum information gain, among others.
	
\subsection{Gradient Boosted Decision Trees}
We provide a brief introduction to XGBoost and the related terms and notations used throughout this paper. We refer the readers to \cite{chen2016xgboost} for further details.
	
Given a dataset $\mathcal{D}$ with $n$ samples and $m$ features, $\mathcal{D}=\{(x_i, y_i)\}^{n}_{i=1}$, where $x_i \in \mathbb{R}^m$ and $y_i\in\mathbb{R}$, the predictions from the XGBoost model, $\hat{y}_i$, is defined as a tree-based additive ensemble model, $\phi(x_i)$, comprising of $K$ additive functions, $f_k$, defined as:
$$\hat{y}_i = \phi(x_i) = \sum_{k=1}^{K} f_k(x_i), \; f_k \in \mathcal{F}$$
	
where $\mathcal{F} = \{f(x) = w_{q(x)}\}$ is a collection of Classification and Regression Trees (CART), such that $q(x)$ maps each input feature $x$ to one of $T$ leaves in the tree by a weight vector, $w \in \mathbb{R}^T$.
	
Given the function defined above, the XGBoost algorithm minimizes the following regularized objective function:
$$\tilde{\mathcal{L}}=\sum_i l(y_i,\hat{y}_i) + \sum_k \Omega(f_k)$$
	
\noindent where $l(y_i,\hat{y}_i)$ is the loss function of the $i$th sample between the prediction $\hat{y}_i$ and the target value $y_i$, and $\Omega(f_k)=\gamma T + \frac{1}{2}\lambda\|w\|^2$ is the regularization component. This component discourages each $k^{th}$ tree, $f_k$, from over-fitting through hyperparameters  $\lambda$, the regularization parameter penalizing the weight vector $w$, and $\gamma$, a term penalizing the tree from growing additional leaves.
	
To approximate the loss function, a second-order Taylor expansion function is used:
$$\mathcal{L}^{(t)} \simeq \sum_{i=1}^n [l(y_i,\hat{y_i}^{(t-1)})+g_if_t(x_i)+\frac{1}{2}h_if_t^2(x_i)] + \Omega(f_t)$$
	
\noindent As the tree is trained in an additive manner, each iteration index of the training process is denoted as $t$, hence $\mathcal{L}^{(t)}$ denotes the $t^{th}$ loss of the training process. Here, we also define the gradient, $g_i = \partial_{\hat{y}_i^{(t-1)}}l(y_i, \hat{y}_i^{(t-1)})$, and hessian, $h_i = \partial_{\hat{y}_i^{(t-1)}}^2 l(y_i, \hat{y}_i^{(t-1)})$ based on the defined loss function. 
	
Given the derived gradients and hessians for a given $q(x)$, we can compute the optimal weights of leaf $j$ using:
$$w^*_j = -\frac{G_j}{H_j + \lambda}$$
	
\noindent where $G_j = \sum_{i \in I_j} g_i$ and $H_j = \sum_{i \in I_j} h_i$ are the total summation of the gradients and hessians for each of the specific data sample indices, $I_j$, respectively. However, to efficiently compute the optimal weights $w^*_j$, we can greedily maximize the following gain score to search for best split for a leaf node at each iteration efficiently by:
$$Gain=\frac{1}{2}\left[\frac{G^2_L}{H_L + \lambda} + \frac{G^2_R}{H_R + \lambda} - \frac{(G_L+G_R)^2}{(H_L+H_R)+\lambda} \right] - \gamma$$
	
\noindent Here, $L$ and $R$ correspondingly considers the sum of the gradients and hessians based on the specific index of the left and right children of the given leaf node, $I_L$ and $I_R$, respectively.

As the ensemble tree model is optimized through an iterative process over additive functions, and not over a euclidean space, the resulting implementation within a FL setting is non-trivial. As the weight vector, $w$, in this model corresponds to specific leaves which are generated with respect to the local parties' data, traditional model exchange techniques such as FedAvg cannot be applied to combine the model parameters at the aggregator. That is, we cannot locally train independent XGBoost models at each parties and easily consolidate their resulting tree structures at the aggregator. In the next section, we describe our method to address how we overcome this limitation through our proposed approach.
\section{Party Adaptive XGBoost}\label{sec:pax}
In this section, we describe our approach, the \textit{Party-Adaptive XGBoost} (PAX) for training a federated histogram-based gradient boosted tree. 
First, we describe the key challenges and intuition behind our method. Then, we present our proposed method as depicted in Algorithm \ref{alg:pax_algo}.
	
\subsection{Challenges and intuition}
The fundamental problem in training a gradient boosted decision tree is to find the optimal feature and value to split using the computed gain score, as described in the previous section. This entails that for each data sample and feature, we compute the gradients, $g_i$ and hessians, $h_i$, of the model to derive a gain score. 
Most of the FL-based gradient boosting algorithms allow each party to transmit the gradients, hessians, and/or feature-value split candidates to the aggregator. There are multiple ways of determining the optimal split in FL.
	
One method in searching for the optimal split uses the \textit{exact greedy algorithm}, as referred to by \cite{chen2016xgboost}, where the method enumerates over the entire space of features and values to find the optimal split \cite{pedregosa2011scikit,ridgeway2006gbm}. However, doing so would not only incur a huge computational cost, but also communication costs over a network of distributed FL parties.
	
To mitigate this performance issue, algorithms such as XGBoost \cite{chen2016xgboost} and LightGBM \cite{ke2017lightgbm} utilize a \textit{quantile approximation strategy} to reduce the overall search space of the split finding procedure as a histogram representation of the raw data distribution. Empirically, quantile approximations of data can provide optimal splits just as well as the exact greedy solution \cite{chen2016xgboost,ke2017lightgbm,keck2017fastbdt}. One observation here is that by quantizing or reducing the resolution of the raw data distribution, we can generate a surrogate representation of the raw data.
However, applying this directly into FL settings may result in models that do not adapt to the deviations of each party's data.
	
To overcome this difficulty, instead of homogeneously approximating the data across all parties in the same manner, our approach takes into consideration the relative ratio of the size of the party's local dataset with respect to those of the other parties in the FL network to determine the appropriate policy for discretizing the data. 
We can achieve this by adjusting the quantile sketch approximation error hyperparameter $\epsilon$\footnote{Note: While $\epsilon$ is used here as a notation for the histogram approximation error, this term does not collude with the notation used in differential privacy (DP).}, used by XGBoost for \textit{each individual party}. Intuitively, the inverse of the $\epsilon$ parameter, or $1 / \epsilon$, is roughly equivalent to the \textit{bin size of the histogram}. Therefore, this approximated surrogate representation of the party's data can be transmitted to the aggregator without directly exposing the \textit{precise} representation of the party's local raw dataset. 
	
Given that each party contributes a histogram of different bin sizes, another key challenge we encounter is how we can best fuse together the histograms. By considering our target loss objective, we can choose the best bin size to align our histogram on to ensure that we can obtain the most accurate representation to find the best split value candidate on.
	
\subsection{PAX}
\begin{algorithm}[!h]
	\DontPrintSemicolon
	\caption{Party Adaptive XGBoost (PAX)} 
	\label{alg:pax_algo}
		
	\KwInput{$\mathcal{D}$, Input Dataset; $A$, Aggregator; $P$, Participating Parties in FL Training; $\mathcal{\epsilon^{(A)}}$, Global Error Tolerance Budget; $T$, Maximum Number of Training Rounds; $l$, Model Loss Function}
		
	\KwOutput{$f^{(\mathcal{A})}$, Trained Global XGBoost Model}
		
	$\mathcal{A}$ Initialize Global Null Model: $f^{(A)}_\emptyset \gets 0$
		
	$\{\bm{\epsilon_1}, ..., \bm{\epsilon_{|\mathcal{P}|}}\} \gets compute\_local\_epsilon(\epsilon^{(\mathcal{A})})$
		
	\;
		
	\For{$i=1, ..., |\mathcal{P}|$}{
		$\mathcal{A}$ Transmits $\bm{\epsilon_i}$ to $p_i$
			
		Assign Local Epsilon $\bm{\epsilon_i}$ to $p_i$
			
		$\tilde{\mathcal{D}}^{(p_i)} \gets compute\_histogram(\mathcal{D}^{(p_i)}, \bm{\epsilon_i})$
	}
		
	\;
		
	\Repeat{$t \leq T$ or other termination criteria.}{
		($\bar{\mathcal{D}}^{(\mathcal{A})}, G^{(\mathcal{A})}, H^{(\mathcal{A})}) \gets (\emptyset,  \emptyset,  \emptyset$)
			
		\For{$i=1, ..., |\mathcal{P}|$}{
			$\mathcal{A}$ Transmits $f^{(\mathcal{A})}_t$ to Party: $f^{(p_i)}_t \gets f^{(\mathcal{A})}_t$
				
			$p_i$ Generate Predictions: $\hat{y}_t^{(p_i)} = f_t^{(p_i)}(\tilde{\mathcal{D}}^{(p_i)}_X)$
				
			$p_i$ Computes $g^{(p_i)}$ and $h^{(p_i)}$
				
			$p_i$ Transmits $g^{(p_i)}$, $h^{(p_i)}$, and $\tilde{\mathcal{D}}^{(p_i)}_X$ to $\mathcal{A}$
				
			$\bar{\mathcal{D}}^{(\mathcal{A})} \gets \bar{\mathcal{D}}^{(\mathcal{A})} \cup \tilde{\mathcal{D}}^{(p_i)}_X$
				
			$G^{(\mathcal{A})} \gets G^{(\mathcal{A})} \cup g^{(p_i)}$
				
			$H^{(\mathcal{A})} \gets H^{(\mathcal{A})} \cup h^{(p_i)}$
		}
		\uIf{$l$ is a classification objective}{
			$\epsilon^{(\mathcal{A})}_m \gets min(\{\bm{\epsilon_1}, ..., \bm{\epsilon_{|\mathcal{P}|}}\} )$
		}
		\uElseIf{$l$ is a regression objective}{
			$\epsilon^{(\mathcal{A})}_m \gets max(\{\bm{\epsilon_1}, ..., \bm{\epsilon_{|\mathcal{P}|}}\} )$
		}
		$\bar{\mathcal{D}}^{(\mathcal{A})}_m, G^{(\mathcal{A})}_m, H^{(\mathcal{A})}_m \gets merge\_hist(\bar{\mathcal{D}}^{(\mathcal{A})}, G^{(\mathcal{A})}, H^{(\mathcal{A})}, \epsilon^{(\mathcal{A})}_m)$
			
		$f_t^{(\mathcal{A})} \gets grow\_tree(\bar{\mathcal{D}}^{(\mathcal{A})}_m, G^{(\mathcal{A})}_m, H^{(\mathcal{A})}_m)$
	}
		
	\;
		
	\SetKwProg{Fn}{Function}{}{end}
	\Fn{compute\_local\_epsilon($\epsilon^{(\mathcal{A})}$): }{
		$S \gets \emptyset$
			
		\For{$p_i \in \mathcal{P}$}{
			$\mathcal{A}$ Queries Data Count: $S \gets S \cup |\mathcal{D}^{(p_i)}|$
		}
			
		$E \gets \emptyset$
			
		\For{$i=1, ..., |\mathcal{P}|$}{
			Compute $i^{th}$ Party $\epsilon$: $\bm{\epsilon_i} \gets \epsilon^{(A)} \left( \frac{s_i}{\sum_{q\in\mathcal{S}}q} \right)$
				
			$E \gets E \cup \bm{\epsilon_i}$
			}
		return $E$}
\end{algorithm}
	
To address the challenges, our proposed method, PAX, computes a party-specific approximation factor which is used to balance the overall information gain from each party while maintaining the minimum necessary granularity of the information they would provide to the global model. 
To place this into the FL context, our method is dependent on: i) how we determine and construct the best surrogate histogram representation for each party and ii) how the histograms are aggregated from each party and utilized within the downstream modeling task for the gradient boosting decision tree.
	
To train a gradient boosted decision tree within a FL-setting, the aggregator first initializes a global null model, $f_{\emptyset}^{(\mathcal{A})}$ (line 1). The aggregator also defines the global hyperparameter $\epsilon^{(A)}$, which denotes the error tolerance budget for the training process for XGBoost. This parameter sets the upper-bound histogram approximation error and equivalently the number of maximum bins used in training.
	
Given the defined global $\epsilon^{(A)}$ parameter, we then determine the appropriate policies for how we construct the local party's surrogate histogram by computing the party's local $\bm{\epsilon_i}$ parameter (line 2). The aggregator first queries each party for their dataset size, $|d_i|$, and maintains a list of counts, which we denote here as $S$. For each $i^{th}$ party, we compute the local party's $\bm{\epsilon_i}$ parameter (line 36):
$$\bm{\epsilon_i} = \epsilon^{(A)} \left( \frac{|d_i|}{\sum_{d\in\mathcal{D}}|d|} \right)$$
	
\noindent Equivalently, rewriting the inverse term of $\bm{\epsilon_i}$ can give us the corresponding number of bins that each $i^{th}$ party will use to construct their surrogate histogram representation of the local dataset:
$$\frac{1}{\bm{\epsilon_i}} = \frac{1}{\epsilon^{(A)} \left( \frac{|d_i|}{\sum_{d\in\mathcal{D}}|d|} \right)} = \frac{\sum_{d\in\mathcal{D}}|d|}{\epsilon^{(A)}|d_i|}$$

The aggregator replies to each party with their respective $\bm{\epsilon_i}$ parameter, and correspondingly assigns the local party epsilon (line 5-6). Each party constructs the surrogate histogram representation of the local party's raw data distribution using methods such as the GKMethod \cite{greenwald2001space}, the extended GKMethod \cite{zhang2007fast}, DDSketch \cite{masson2019ddsketch}, and other viable weighted quantile sketch method. For this purpose, the party uses the given $\bm{\epsilon_i}$ and their respective local data distribution, $\mathcal{D}^{(p_i)}$ to compute a surrogate histogram representation of the local party's data distribution, $\tilde{\mathcal{D}}^{(p_i)}$ (line 7).
	
After computing the surrogate histogram representation of the data, we then initiate the iterative federated learning process. First, the aggregator $\mathcal{A}$ transmits their global model, $f^{(\mathcal{A})}_t$ to each party, $p_i$, which is assigned to each party's respective local model $f^{(p_i)}_t$ (line 13). We then evaluate $f_t^{(p_i)}$ on $\tilde{\mathcal{D}}^{(p_i)}_X$, the input feature histogram data, $X$, with respect to the $i$th party's training data, $p_i$, to obtain the model's predictions, $\hat{y}_t^{(p_i)}$ (line 14). Afterwards, given the predictions, we compute the loss function which is used to compute the gradient, $g^{(p_i)}$, and hessians, $h^{(p_i)}$ for each of the corresponding surrogate input feature value split candidates (line 15). Gradient and hessian statistics that fall under a certain bin interval are grouped together within their respective value buckets \cite{chen2016xgboost}. These values are then sent back as replies to the aggregator and collected until all parties have finished computing these three values (line 16).
	
Given the collected results from each party, we perform a fusion operation to merge the final histogram representation used towards boosting the decision tree model. This entails reconstructing the local party's histogram by re-aligning the bin size from each of the collected histograms accordingly to a particular $\epsilon$ value. This is determined by a heuristic based on the defined downstream objective loss function, $l$ (line 21-24). If the loss function task is based on a classification loss such as binary cross-entropy or categorical cross-entropy loss, we consider the smallest $\epsilon$ (\textit{i.e. large bin size}) computed from each of the local party's data distribution. Conversely, if the task is based on a regression based loss function, such as the squared loss function or root mean squared error (RMSE), then we consider a large $\epsilon$ (\textit{i.e. small bin size}). Here the primary focus of the heuristic is based on the cardinality and resolution of the target label data.
	
The final $\epsilon$ value based on this heuristic is denoted as $\epsilon^{(\mathcal{A})}_m$, where $m$ denotes the variables pertaining to the merging process. Using the derived $\epsilon^{(\mathcal{A})}_m$, we can utilize existing methods such as \cite{chenxgboost} and \cite{blomer2015large} for implementing merging histogram routines with error guarantees based on some defined error parameter. This produces the final outputs for the merged gradients, $G^{(\mathcal{A})}_m$,  hessians, $H^{(\mathcal{A})}_m$, and feature-value split candidates, $\bar{\mathcal{D}}^{(\mathcal{A})}_m$ which is used for the boosting process (line 25-26). With a new $f_t^{(\mathcal{A})}$ generated, we repeat our training process for $T$ rounds, or until some other stopping criteria depending on whether early-stopping or other heuristics are considered (line 27).
\section{Experiments} \label{sec:numerical}
In this section, we describe two experiments to evaluate the efficacy of PAX's performance and robustness. The first set of experiment is based on a model evaluation, where we train our model and benchmark our results against other FL algorithms to train XGBoost. The second experiment systematically explores the sample distribution balance of the dataset and how this heterogeneity can influence the overall aggregation performance of the trained models.
	
\subsection{Model Evaluation}
\subsubsection{Experimental Setup}
We evaluated our models based on a three-party FL setup, that was trained primarily on two FL frameworks which were executed on a simulated FL environment (i.e. not run on the real distributed environment). For this, we do not consider network latency issues within the scope of our evaluation. All experiments were executed on a 2018 MacBook Pro with a 2.6 GHz 6-Core Intel Core i7 CPU and 32 GB of memory. 
	
\subsubsection{Models}
In our evaluation, we trained three different FL models: a) our Party Adaptive XGBoost (PAX) algorithm, b) Homo Secureboost, a horizontal variant of the gradient boosting decision tree model implementation  from FATE \cite{fate:2020}, and c) a FL-based implementation of the logistic regression model based on the weight averaging aggregation method. For both gradient boosting models, the key parameters we have defined are the global bin size to be at 255 and the maximum rounds to be at 100.
	
\subsubsection{Dataset and Evaluation Methodology}
We evaluate our models on the airline delay causes dataset from the U.S. Department of Transportation's (DOT) Bureau of Transportation Statistics (BTS) \cite{dot2009}. We use a version of the dataset that was prepared and preprocessed from Kaggle \cite{kaggle2020}. The dataset features 30 attributes pertaining to flight times, status, and delay information with many of the features being significantly skewed and non-IID. In our modeling task, we considered a binary classification where our target attribute is to predict whether or not a flight would be delayed or not. Each party is trained on 1,000 samples, based on two scenarios where the samples were selected randomly (based on uniform distribution) and another variant of the dataset where the samples selected are balanced based on the target labels. For testing set of our model, we sample another 1,000 instances based on the two different sampling methods. For our evaluation metrics, we consider accuracy (ACC), precision (PRE), recall (REC), area under the curve (AUC), and the F1 score (F1). Our reported results will be based on taking the average of the three parties' metrics based on the evaluation on the test set. Furthermore, for each algorithm, we also report the total execution time for training the model.
	
\subsection{Data Partition Evaluation to Emulate Party's Data Heterogeneity}
\begin{table}[!htbp]
	\small
	\centering
	\begin{tabular}{c|ccc}
	    \textbf{Partition} & \textbf{Party 1} & \textbf{Party 2} & \textbf{Party 3} \\ \hline
		\textbf{1}         & 1000 (33\%)      & 1000 (33\%)      & 1000 (33\%)      \\
		\textbf{2}         & 1500 (50\%)      & 1160 (39\%)      & 340 (11\%)       \\
		\textbf{3}         & 2080 (69\%)      & 804 (27\%)       & 116 (4\%)        \\
		\textbf{4}         & 2482 (83\%)      & 478 (16\%)       & 39 (1\%)         \\
		\textbf{5}         & 2721 (91\%)      & 265 (9.6\%)      & 13 (0.4\%)       \\ \hline
	\end{tabular}
	\caption{Per-Party Data Sample Partition Sizes}
	\label{tab:data_partition}
\end{table}
To evaluate the robustness of our method, we perform a systematic analysis of our algorithm's performance by re-partitioning the data samples across the three parties. The main motivation behind this analysis is to evaluate how well our algorithm can adapt to different dataset sizes across each party and how they can influence the average performance of our model. For this evaluation, we utilize the same randomly sampled airline delay dataset from our previous evaluation and reallocate the distribution of the data. In the first allocation (Partition 1), we begin with each party having the same number of samples. From each partition types, we re-partition our dataset by first randomly taking half (50\%) of the data from Party 2 and appending it to Party 1. Then, we similarly take approximately two-thirds of the data from Party 3 and move it to Party 2. The result of this process yields the five different partitions as shown in Table \ref{tab:data_partition}.
	
\subsection{Experiment Results}
\begin{table}[h]
	\tiny
	\centering
	\resizebox{\textwidth}{!}{%
		\begin{tabular}{c|ccccc|ccccc}
			\hline
			\multicolumn{1}{l|}{} & \multicolumn{5}{c|}{\textbf{Airline (Random)}}                                      & \multicolumn{5}{c}{\textbf{Airline (Balanced)}}                                         \\ \hline
			\textbf{Model}        & \textbf{ACC}  & \textbf{PRE}  & \textbf{REC}  & \textbf{AUC}  & \textbf{F1}   & \textbf{ACC}  & \textbf{PRE}  & \textbf{REC}  & \textbf{AUC}  & \textbf{F1}   \\ \hline
			\textbf{PAX (Ours)}   & \textbf{0.88} & \textbf{0.88} & \textbf{0.88} & \textbf{0.87} & \textbf{0.88} & \textbf{0.87} & \textbf{0.85} & \textbf{0.90} & \textbf{0.87} & \textbf{0.88} \\
			Homo SecureBoost           & 0.74          & 0.71          & 0.82          & 0.81          & 0.72          & 0.81          & 0.84          & 0.79          & 0.86          & 0.81          \\
			Logistic Regression   & 0.58          & 0.72          & 0.49          & 0.58          & 0.45          & 0.58          & 0.72          & 0.49          & 0.58          & 0.45          \\ \hline
		\end{tabular}%
	}
	\caption{Results for Model Evaluation on Testing Samples for Airline Random \& Balanced}
	\label{tab:model_eval}
\end{table}

In this section, we present our results for each of the experiments described above. Table \ref{tab:model_eval} demonstrates our results from each of the three models and the two other XGBoost evaluations. For each dataset variation, Airline Random and Balanced, we report the metrics described in our experimental setup. From our experimental results, we can empirically observe that our model outperforms both the Homo SecureBoost and the Logistic Regression models achieving 88\% and 87\% accuracy for the random and balanced datasets respectively. Despite significant improvements for the remaining two models evaluated under the balanced dataset, we observe that our approach still outperforms over both methods. From these results, we can infer that majority of the errors from Homo SecureBoost derive from the randomly appended number used in their secure histogram aggregation method, which impacted the overall choice of the feature-value threshold during the boosting process, therefore yielding lower accuracies compared to our approach. 

\begin{table*}[!htbp]
	\normalsize
	\centering
	\begin{tabular}{cccc}
		& \textbf{PAX} & \textbf{SecBoost} & \textbf{LogReg} \\ \hline
		\multicolumn{1}{c|}{Airline (Random)}   & 90 sec       & 1172 sec             & 15 sec                       \\
		\multicolumn{1}{c|}{Airline (Balanced)} & 49 sec       & 1177 sec             & 16 sec                       \\ \hline
	\end{tabular}
	\caption{Model Training Walltime Comparison}
	\label{tab:walltime}
\end{table*}

Table \ref{tab:walltime} shows our training runtimes for each of the respective algorithms evaluated on each of the datasets. Comparing between our method and SecureBoost, our model takes significantly less time to train compared to Homo SecureBoost. As FATE's Homo SecureBoost utilizes a cryptographic routines for data protection, this consequently incurs even more computational overhead as exhibited in the walltime measurements obtained in our experiments. On the other hand, although the runtime for logistic regression is significantly much shorter than that of our method, the time taken does not fully justify for the complexity of the model and the poor results obtained in the modeling task (Table \ref{tab:model_eval}).

\begin{table}[!htbp]
	\centering
	\begin{tabular}{c|ccccc}
		\multicolumn{1}{l|}{\textbf{Partition}} & \textbf{ACC} & \textbf{PRE} & \textbf{REC} & \textbf{AUC} & \textbf{F1} \\ \hline
		\textbf{1}                              & 0.88         & 0.88         & 0.88         & 0.87         & 0.88        \\
		\textbf{2}                              & 0.79         & 0.74         & 0.87         & 0.79         & 0.80        \\
		\textbf{3}                              & 0.84         & 0.81         & 0.88         & 0.85         & 0.84        \\
		\textbf{4}                              & 0.92         & 0.90         & 0.93         & 0.92         & 0.92        \\
		\textbf{5}                              & 0.91         & 0.89         & 0.91         & 0.91         & 0.91        \\ \hline
	\end{tabular}
	\caption{Dataset Partition Experiment Results}
	\label{tab:partition_results}
\end{table}
	
Table \ref{tab:partition_results} shows our empirical results obtained from the data partition evaluations that analyze the performance under substantially different levels of data heterogeneity among different parties. Overall, we see that despite how we change the partitions, PAX is able to maintain robust performance in shifting distributions. Even in extreme partition scenarios, where Party 1 held over 90\% and Party 3 held less than 1\% of the data (Partition 5), we see that our method even improved performance over the balanced results achieving the highest accuracy of 91\% - even greater than when the dataset is balanced (Partition 1). We speculate that this increase in accuracy occurred as Party 1 holds the majority of the data in the federation, thereby having enough samples to be able to provide a sufficient amount of information without having to compensate a huge loss of accuracy. Conversely, Party 3 only having few samples would be a major risk to exposing raw data does not provide as much to the model, and therefore contributes a relatively small amount of information to the modeling process. This empirically demonstrates that our proposed method is robust as it is able to adapt to different shifts in the data distribution and adopt the best histogram based on the learning task at hand. Further, these results show how PAX overcomes one of the main challenges of training models in FL.
\section{Conclusion}\label{sec:conclusion}
In this paper, we proposed the Party-Adaptive XGBoost (PAX) approach for training a gradient boosted decision tree model in a Federated Learning setting. In this approach we use histograms to communicate from parties to the aggregator. PAX surrogate histograms can be adapted to the specifics of the properties of a party’s data and the downstream learning task and, thus, can address imbalances of data between parties, which are commonly difficult to deal with in a federated learning setting. In both our empirical model evaluation and partition reallocation experiments, we have demonstrated  that our method is able to outperform state of the art methods both in model performance and training time. This applies to the benchmark Homo Secure Boost approach as well as to the federated logistic regression we use as a reference point. The good model performance and overall faster training time, our method makes it suitable for enterprise federated learning use cases not requiring consumer use case differential privacy guarantees.

\bibliographystyle{unsrt}
\bibliography{References}

\end{document}